# Architecture of Automated Crypto-Finance Agent

Ali Raheman
*Autonio Foundation Ltd.*
Bristol, UK
ali.raheman@gmail.com

Anton Kolonin
*SingularityNET Foundation*
Amsterdam, Netherlands
akolonin@gmail.com

Ben Goertzel
*SingDAO Ltd.*
Gros-Islet, St.Lucia
ben@goertzel.org

Gergely Hegyközi
*Autonio Foundation Ltd.*
Bristol, UK
gergely.hegykozi@gmail.com

Ikram Ansari
*Autonio Foundation Ltd*
Bristol, UK
ikram.ansari101@gmail.com

*Abstract*—We present the cognitive architecture of an autonomous agent for active portfolio management in decentralized finance, involving activities such as asset selection, portfolio balancing, liquidity provision and trading. Partial implementation of the architecture is provided and supplied with preliminary results and conclusions.

## I. Introduction

The subject of decentralized finance is attracting the attention of investors as well developers and scientists due to the high potential financial returns, increased demand for implementation of automated business applications for investments, liquidity provision, and trading using crypto-currencies. Enormous volatility and the presence of "on-chain" data such as transaction logs that may be used as an extra source of data for applications based on artificial intelligence and machine learning are a few unique properties of crypto-financial markets.

The key possibility associated with decentralized finance is automated liquidity provision, called market making, which can be performed on either centralized exchanges (CEX) such as Binance, or decentralized ones (DEX) such as the smart contracts Uniswap or Balancer on the Ethereum blockchain. How machine learning and artificial intelligence can be applied, attempting to learn efficient market making strategies, is a matter of active study: [1,2,3,4]. So far, the results are not that exciting: demonstrated ability to learn some basic principles of trading using limit book orders with the ability to outperform "hodling" strategy (buy and hold on rising market) in very specific conditions. So more effort is required to succeed in this area.

The important part of automated trading is a price prediction [5,6] which can take the form of either predicting price change direction as a classification problem or prediction of specific price level as a regression problem. While the latter appear more critical for market making activity, that is because conventional trading with market orders could accept predicted price direction change as a trading signal for either sell or buy. In turn, market making with limit book orders on a CEX or swap pools on a DEX don't need to sell or buy, they just need to set the appropriate price levels on bid and ask orders on a CEX or adjust the pricing function on a DEX according to anticipated price movement and the actual target level of its move. Unfortunately, the high volatility and manipulative nature of the crypto market provides challenges for the former and latter.

Finally, asset selection and portfolio balancing and rebalancing according to market dynamics remains a critical activity to active portfolio management, including the crypto markets. That requires appropriate metrics to evaluate assets for both inclusion of them in portfolios as well as re-weighting the portfolios as time goes on and the market changes. The volatile nature of the crypto markets provide challenges for some of the traditional metrics like "Sharpe Ratio" which can not be allied on "bear" markets. That involves the need to search for advanced ways of asset quality assessment [7], such as "Modified Sharpe Ratio" [8,9]. Moreover, for the purpose of active portfolio management involving active trading of the vested assets by portfolio manager it might be not applicable either as the variance reduces the quality of an asset according to most of the conventional metrics, while the trading strategy exploiting variance may actually be granting more value for a volatile asset.

The presented work will focus on the approach to deal with the above mentioned challenges and provide some preliminary results.

## II. Overall Architecture

### A. System Components and Layers

Below we provide an overall design infrastructure involving assembly of artificial intelligence (AI) agents, or "Oracles," for use across generic portfolio management, liquidity provision and price prediction prices and on various decentralized financial markets. We expect to deploy an ecosystem of such AI Oracles to support investment decisions on a platform hosting the Oracles to help increase the value and returns of the investments by means of providing liquidity to decentralized cryptocurrency markets. Each of the agents in the ecosystem will be serving as an AI Oracle for end business applications, smart contracts, and other agents. The scope of business activities to be served by the AI Oracles can be seen on the following diagram. The important part of the diagram is that it assumes explicit difference between "inventory/portfolio" (in the scope of "Portfolio Balancer"), which may aggregate multiple assets/instruments for execution and the "DEX swap pool" or "DEX balancing pool" (in the scope of Market Maker/Trader) which may be just one of the strategies involved in maintenance of the portfolio (other strategies may be hodling, trading on a DEX or CEX, providing limit orders on CEX, etc.). That means, one "inventory/portfolio" may have multiple "DEX swap pools" or "DEX balancing pools" with different strategies, weighings, pricing curves, etc.

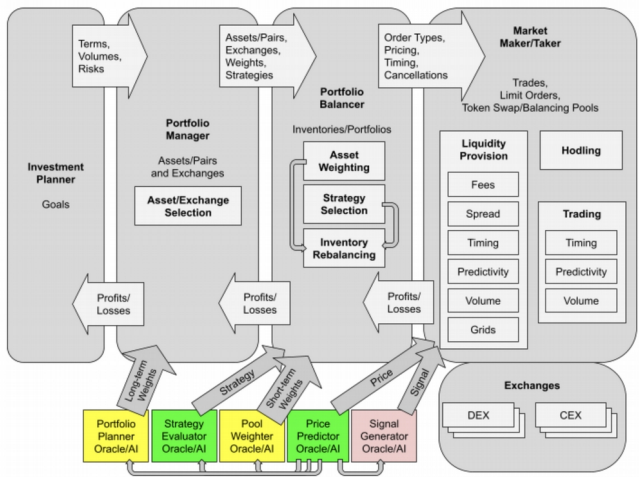

Fig.1. Business functions/applications (at the top) served by AI Oracles (at the bottom).

How the AI oracles interact with each other and the data sources may be seen on the following diagram.

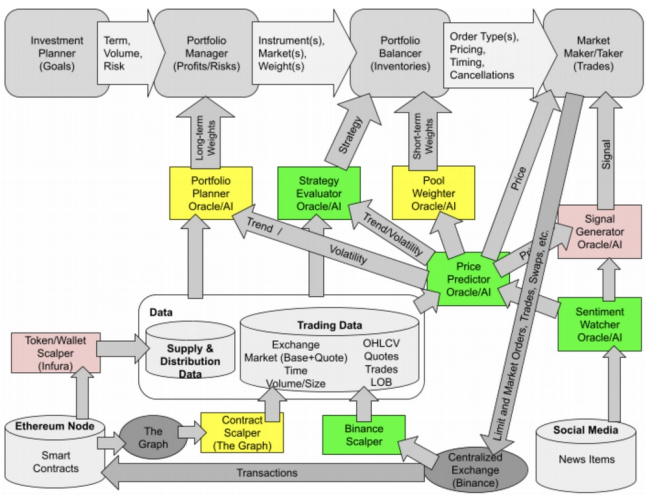

Fig.2. AI Oracles (in the middle) serving business functions/applications (at the top) relying on data scalping services (at the bottom).

The **Portfolio Planner Oracle/AI** agent will be using accumulated on-chain market data as well as getting predictions on price trends and volatility from the Price Predictor Oracle/AI to provide long-term weights on market instruments (tokens); therefore, helping human or programmatic Portfolio Managers building long-term investment portfolios («DynaSets») given the specified terms, volumes and acceptable levels of risk.

The **Strategy Evaluator Oracle/AI** will be using the same on-chain data and price trend/volatility predictions to evaluate different competitive strategies and parameters of these strategies using backtesting on historical data so the winning strategy corresponding to current market conditions could be recommended for current operations on a portfolio maintained by Portfolio Balancer applications and smart contracts: rebalancing its inventory, deploying smart contracts for liquidity provision on the portfolio instruments and executing corresponding trades.

The **Pool Weighting Oracle/AI** will be relying on the same data and predictions suggesting short-term weights on market instruments (tokens) helping Portfolio Balancer to adjust portfolio inventory given short-term risks.

The **Signal Generator Oracle AI** will be taking predictions of current price fluctuations and sentiment buzz in respect to specific instruments (tokens) to generate signals for trading and liquidity provision applications and smart contracts for when to buy, sell, create or cancel limit orders as well as the optimal sell, buy, ask and bid prices appropriate for that given the market momentum.

The **Sentiment Watcher Oracle/AI** will be monitoring news feeds on social and online media in respect to specific instruments (tokens) and overall crypto-related buzz to provide overall assessment of the sentiment for Signal Generator and Price Predictor for their inferences.

The Price Predictor appears to be a key component of the ecosystem serving predictions of price trends and volatility at different prediction horizons for all multiple other agents listed above. It uses AI/ML relying on all sorts of market, fundamental, and sentiment data available to serve the rest of the ecosystem with predictions based on that.

The data feeding the AI Oracle agents will be collected in a number of ways: scalping centralized exchanges, such as Binance, for trading data, trades, and limit order book snapshots (Binance Scalper), crawling DEX on-chain trades on protocols such as Uniswap V2 using 3rd party data gateways such as The Graph service (Contract Scalper), or monitoring corresponding DEX smart contracts on the live Ethereum Node using Infura API. The information about the overall supply and distribution of the instruments (tokens) will be collected from live Ethereum Nodes as well (Token/Wallet Scalper).

The presented architecture is currently being implemented. Primarily, the Strategy Evaluator, Price Predictor, Portfolio Planner and Pool Weighter are under construction with some of the preliminary results presented further in this paper.

*B. Simulation and Backtesting Architecture in Strategy Evaluator*

The key component of the **Strategy Evaluator** is expected to be Simulation and Backtesting frameworks serving two different yet complementary purposes.

The **Simulation** framework is intended to simulate multi-agent trading and market making activity within a configured environment of virtual agents and market conditions. Given the tentative price curve as an externally driven "fundamental" trend, these agents are interacting on the simulated market either as liquidity providers (LP) posting bid and ask orders on the limit order book or as liquidity takers (LT) executing market orders against the order book. While both LP and LT agents may be following different strategies, they are given the same conditions at the startup of the simulation – in terms of either fixed income rate or initial credit in either base or quote currency. The results are recorded during the simulation and returns or losses of every agent are evaluated at the end.

The **Backtesting** framework is similar to Simulation one, although the price trend is not simulated, but rather taken from real live or historical data. In addition, only a limited set of agents are being simulated, while the activity from other agents is assumed presented by real historical trades and limit order book snapshots. The execution of limit book orders made by simulated data is done solely on real historical or live orders, with account to time, so if no appropriate "real" trade is found matching the time and price of "simulated" limit order, the latter is not executed.

Both frameworks have their pros and cons. The Simulation may be oversimplifying behavior of a real market, removing actual agents executing the strategies not anticipated by the simulated population, but it can actually render the "zero-sum game" [10] phenomenon known to low liquidity markets such as crypto markets. The Backtesting may be used to evaluate the strategies being prospected on the real market patterns while the "zero-sum game" effects will be undercounted because no actual impact on the market will be happening due to activity of the "simulated" agents.

Both frameworks, depending on the presence or absence of the historical or real data, can be used by Strategy Evaluator to assess applicability of one or another market making or taking strategy given the inventory of an assets in prospective or actual portfolio under target market conditions.

*C. Price Predictor*

The Prediction Oracle (Price Predictor) is supposed to take a key role in the suite of AI agents so that thePrice Predictor provides the following inputs for the other services and AI oracles, as shown in the figures above. Expectation of price for specific time in the nearest future can be used by Market Makers/Takers to execute their trading (market or limit) orders. Expectations of price trends and volatility levels can be used by Portfolio Balancer and Portfolio Manager for the purpose of inventory re-weighing (Portfolio Manager) or selection of the strategy for inventory rebalancing (Portfolio Balancer).

The main principle of the Predictor is the ability to perform predictions for "expected market price" values for target symbol pairs ("markets") on specific "exchanges" in real time: being able to update its model on the fly using long-term historical data and provide predictions at the same time using short term live data. The Predictor can simultaneously do both at the same time, around the clock, potentially applying them to different symbol pairs and exchanges – see the following figure for explanation.

**Incremental training.** Update predictive models by re-training on historical data spanned over temporal "training interval" (hour, day, week, month, quarter, half-year, year) periodically, using a "period" increment (5 minutes, hour, 6 hours, day, week, month, etc.) and "historical interval" window to get the features from the entire "training interval." Respective to the historical intervals, the "period" is somewhat less than the "training interval" and the "training intervals" overlap with increments of the "period." When training is being done, each training iteration involves the chunk of data of "historical interval" as input and the chunk of data to be predicted at once as a "batch," so there may be many iterations with overlapping frames of "historical interval" plus "batch" width within the single "training period."

**Incremental prediction.** Use the latest model (learned on the most recent long-term "training interval" upon completion of the last "period") to predict the "expected market price" during the next current "period." The input data used against the model will be the latest rolling "historical interval" covering part of the latest "training interval" and part of the currently being predicted "period," from the very beginning till the very end of the latter. The predictions within the period may be taking place **on the basis** of "batches", so the same "historical interval" may be used to predict price in several subsequent points of time within a single "frame," so the "historical intervals" overlap with increments of the "batch."

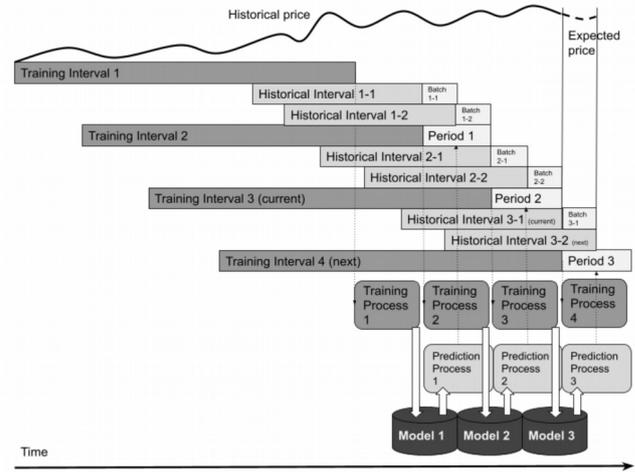

Fig.3. Overlapping training and historical intervals for periodic update of the prediction models, with possibility of batch predictions in real-time environment.

Selection of duration of temporal durations of the "intervals" ("training" and "historical"), "periods," and "batches" may be defined by the runtime performance constraints. Ideally, the "period" and "batch" should both be equal to 1 (minimum time unit such as second) but it might be unrealistic due to run-time performance and response time requirements, such as the following:

- "Training intervals" should be long enough to provide enough data for reliable prediction accuracy but they should also be short enough so the training time takes less than a period.

- "Historical intervals" should be long enough to capture historical data used to derive prediction but short enough so that prediction is done fast enough in order to minimize the size of the "batch."

- "Periods" should be short enough to have the models re-trained on the newest training intervals as often as possible but they should also be long enough to let the training be completed before the end of any period.

- "Batches" should be short enough (ideally - 1 minimum time unit such as second) to let the times of prediction be as close to the current historical interval but long enough to let each prediction act to be completed timely within the duration of the current batch.

- Also, the "batches" may be extended in duration because of the business requirements to have an extended horizon of predictions. In such cases the frames may overlap and have the terminal times of the frames exceeding terminal times of the periods that contain them. Implementation of the batch-based predictions may vary depending on actual prediction algorithms, like follows.
    - If using polynomial extrapolation or linear regression, any number of points can be predicted in one shot.
    - If using neural networks predicting one value at a time, the series of values within the batch window may be predicted iteratively, having each previous value

added to the end of input data for the new position of the historical window.

While the default setup described above assumes that prediction will be always using the very latest model trained on the very latest training interval, there may be different improvements of such principle based on practical considerations such as the following.

- Since some of the symbol pairs ("markets") may be bound to the same fundamental and/or speculative reasons, it could be possible to share/reuse the same model (trained on a single symbol pair) across different symbol pairs. In such a case, a single training process may run on a single historical interval corresponding to the "reference market," creating a "reference model," but multiple prediction processes may run on and apply the same "reference model," with corresponding scaling, to similar "markets" relying on multiple corresponding prediction intervals.

- It might be figured out that models are not specific to the current and the most recent historical data but rather to typical market conditions such as "bullish," "flat," and "bearish" combined with different market caps and types of instruments. In such a case, instead of incrementally re-training the models, it might be better to have a suite of pre-trained models in stock and apply the one that is better matching expected market condition based on separate grand-model matching kind of dynamics of prediction interval and suggesting corresponding "stock model" ("model out of stock" so to speak) instead of training another one.

### D. Portfolio Evaluation Principles in Portfolio Planner and Pool Weighter

For active portfolio management, Portfolio Planner and Pool Weighter have to be evaluated together, with decisions taken in respect to one affecting the decision taken in respect to the other. First, assets included in the portfolio by Portfolio Planner and their weights adjusted by Pool Weighter. Secondly, a strategy to execute the portfolio, such as super-strategy – either "Hodling," "Liquidity Provision" (Market Making), or "Trading" (Liquidity Taking) as indicated on Fig. 1 above. Moreover, there may be more precise identification of each of these super-strategies as a sub-strategy with specific parameters for profit margin, limit order cancellation, order grid settings and many others. In regards to the Pool Weighter it might be not a single strategy, but a combination of the strategies applied to dedicated fractions of an entire portfolio.

In order to evaluate an asset alone, classical approaches derived from return and variance, with "Sharpe Ratio" being the most famous, are questioned if they can be applicable to highly volatile markets [7]. The simplest solution we have found is to use the so-called "Modified Sharpe Ratio" [8,9], which denominates the return by variance in case of positive return but multiples the loss by variance in case of negative return. In this case, if two assets provide the same loss (negative return) for some period of time, the one with smaller variance is ranked as more preferable with overall "Modified Sharpe Ratio" close to zero.

However, for some particular strategies suggested by active portfolio management, referring to variance might be misleading. For instance, if the portfolio manager is performing liquidity provision based on solid fundamental knowledge, semi-insider information or high-quality price prediction models, the volatility might be rather exploited and turned into higher returns than it would be expected based on "hodling."

Because of the latter reason, we anticipate both Portfolio Planner and Pool Weighter to use either Simulation or Backtesting framework to evaluate actual performance of an asset in terms of profits or losses recorded on the basis of actual strategy execution with specific strategy parameters at target market conditions.

### III. PARTIAL IMPLEMENTATION AND INITIAL RESULTS

Given the current state of development, some aspects of the AI Oracles presented above have been tested on real data from Binance CEX.

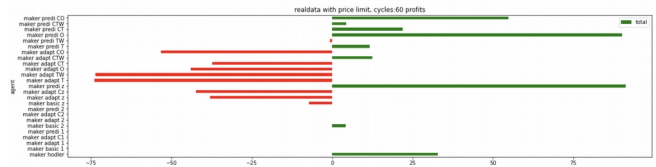

Fig.4. Profits (right bars) and losses (left bars) for market making by different strategies compared to "hodling" (at the bottom) where strategies based on price predictions actually know the future price as if they were having "insider" information.

Simulation and Backtesting have been carried out on Bitcoin BTC/USDT exchange rate on different time intervals for the past half year with consistent results across time intervals and different strategies of market making. It has been shown that almost any selected market making strategy may be profitable (with no losses) if an agent can anticipate the price movements and the price level. In that case, the most profitable strategy has turned to be "zero-spread market making" (in the middle of Fig.4) with ultimate returns compared to "hodling" (at the bottom of Fig.4). The next two strategies using the same "prediction" technology were strategies setting the bid and ask orders at the price level one point better than the competitors – based on the limit order book information known from the last limit order book snapshot (at the top of Fig. 4). All three of the strategies made it possible to outperform the "hodling" strategy substantially.

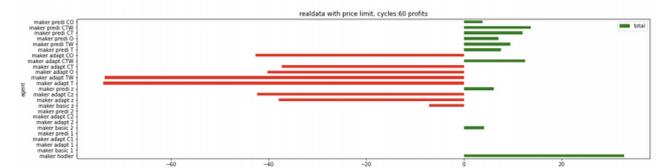

Fig.5. Profits (right bars) and losses (left bars) for market making by different strategies compared to "hodling" (at the bottom) where strategies based on price predictions use linear regression with error level about 10% less than just relying on the last known price. Profits (right bars) and losses (left bars) for market making by different strategies compared to "hodling" (at the bottom) where strategies based on price predictions use linear regression with error level about 10% less than just relying on the last known price.

Advanced experiments with more realistic price prediction technology provided by the Price Predictor discussed above have been run on the same data, as shown on Fig. 5. In fact, at the time of this writing, almost none of the machine learning algorithm known to deal with time series data, including LSTM, Lasso, and Signature, were able to provide accuracy of price level prediction with error less than can be obtained just by looking at the last known price. The only methods improving the results with decrease of error level to 10% less were plain Linear Regression and Ridge Regression. Unfortunately, this level of improvement did not make it possible to outperform the "hodling" strategy.

Portfolio evaluation experiment has been done for five assets from the so-called "Defi-5" (UNI, AAVE, UNI, CRX, COMP) index plus Bitcoin (BTC) and Ethereum (ETH) for specified time intervals during the high volatility of the crypto-market for one hour on May 4, 2021, per minute data sampling. Straight evaluation of the assets with return, variance (stdn), and "Modified Sharpe Ratio" (msharpe) is shown on Fig. 6. For one observation, it is seen that the high volatility of ETH makes it less attractive than AAVE and COMP than it would be based on single return. The latter would be true even for plain "Sharpe Ratio." However, it can also be seen that small variance of SNX and BTC compared to CRV, making them less unattractive in comparison due to the nature of the "Modified Sharpe Ratio."

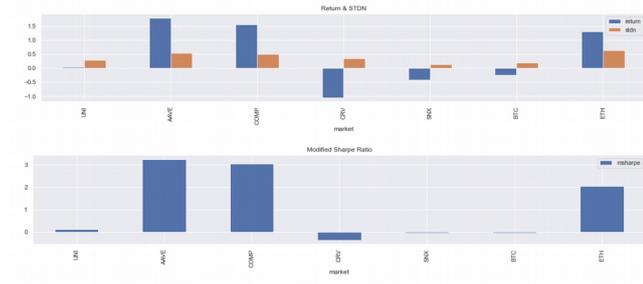

Fig.6. Potentially target assets for portfolio evaluated by return (left bars on the top), variance as normalized standard deviation (right bars at the top) and "Modified Sharpe Ratio" (bars at the bottom).

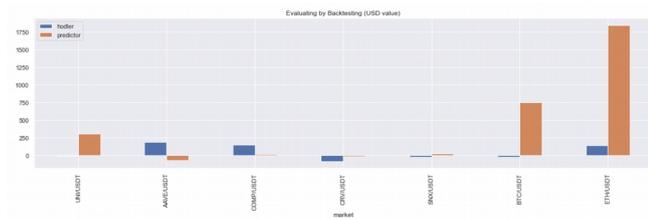

Fig.7. Potentially target assets for portfolio evaluated by backtesting using predictive strategies (right bars) compared to "hodling" (left bars).

Moreover, the same assets were evaluated with backtesting on historical data using the same interval and results have shown quite different distribution of preference. The backtesting was comparing the profits/losses of "hodler" agents (which are proportional to return on the Fig.6) with profits/losses of the agents employing all possible strategies relying on knowledge of the future price. In the latter case, ETH and BTC appears the most preferable, despite negative return (by "hodling") of the BTC and high variance of ETH.

## CONCLUSION

The suggested architecture of an automated agent for active portfolio management in decentralized finance, including portfolio planning and balancing, liquidity provision and trading appear quite flexible, covering all aspects of the crypto-investment business.

The preliminary results show utility of automated simulation and backtesting for strategy selection, value of using contents of the limit order book for liquidity provision on CEX and possibility of increase of the market making profitability even with small increase of accuracy of the price prediction.

The paper is about to be presented at International Symposium on Knowledge, Ontology, and Theory (KNOTH) on 1-5 Dec 2021.

## REFERENCES

[1] S. Ganesh, et. al., "Reinforcement Learning for Market Making in a Multi-agent Dealer Market", arXiv:1911.05892v1 [q-fin.TR] 14 Nov 2019. https://arxiv.org/pdf/1911.05892.pdfGanesh S., et. al. Reinforcement Learning for Market Making in a Multi-agent Dealer Market. arXiv:1911.05892v1 [q-fin.TR] 14 Nov 2019. https://arxiv.org/pdf/1911.05892.pdf

[2] J. Sadighian, "Deep Reinforcement Learning in Cryptocurrency Market Making", arXiv:1911.08647v1 [q-fin.TR] 20 Nov 2019. https://arxiv.org/pdf/1911.08647.pdf

[3] J. Sadighian, "Extending Deep Reinforcement Learning Frameworks in Cryptocurrency Market Making", arXiv:2004.06985v1 [q-fin.TR] 15 Apr 2020. https://arxiv.org/pdf/2004.06985

[4] O. Guéant, et. al., "Dealing with the Inventory Risk. A solution to the market making problem", arXiv:1105.3115 [q-fin.TR] 3 Aur 2012. https://arxiv.org/pdf/1105.3115.pdf

[5] A. Tsantekidis, "Using Deep Learning for price prediction by exploiting stationary limit order book features", arXiv:1810.09965 [cs.LG] 23 Oct 2018 https://arxiv.org/abs/1810.09965

[6] C. Yanjun, et. al., "Financial Trading Strategy System Based on Machine Learning", Hindawi / Mathematical Problems in Engineering Volume 2020, Article ID 3589198, 13 pages, 2020. https://doi.org/10.1155/2020/3589198

[7] H. Scholz, "Refinements to the Sharpe ratio: Comparing alternatives for bear markets", Journal of Asset Management 7(5):347-357 Follow journal DOI: 10.1057/palgrave.jam.2250040, December 2007.

[8] C. Israelsen, "A refinement to the Sharpe ratio and information ratio", Vol. 5, 6, 423–427 Journal of Asset Management, 2005.

[9] J. Alvi, et. al., "Modified Sharpe Ratio Application in Calculation of Mutual Fund Star Ranking", Global Journal of Business Economics and Management Current Issues 10(1):58-82, 2020.

[10] B. Hanley, "A zero-sum monetary system, interest rates, and implications", arXiv:1506.08231 [cs.CE] 26 Jun 2015. https://arxiv.org/pdf/1506.08231.pdf